\documentclass{article}
\usepackage{arxiv}
\usepackage[utf8]{inputenc} 
\usepackage[T1]{fontenc}    
\usepackage{hyperref}       
\usepackage{url}            
\usepackage{booktabs}       
\usepackage{amsfonts}       
\usepackage{nicefrac}       
\usepackage{microtype}      
\usepackage{lipsum}
\usepackage{fancyhdr}       
\usepackage{graphicx}       
\graphicspath{{media/}}     
\usepackage{amsmath}
\usepackage{subcaption}
\usepackage{amssymb}
\usepackage{amsthm}
\usepackage{mathtools}
\usepackage{pdfpages}
\usepackage{float}

\newtheorem{remark}{Remark}

\DeclareUnicodeCharacter{221E}{-}

\usepackage{algorithm}
\usepackage{algpseudocode}
\newcommand\ignore[1]{{}}
\algrenewcommand\algorithmicrequire{\textbf{Input:}}
\algrenewcommand\algorithmicensure{\textbf{Output:}}

\usepackage{caption}

\DeclareMathOperator*{\argmin}{arg\,min}

\title{
Deep Incremental Model Informed Reinforcement Learning for Continuous Robotic Control
}

\author{
  Cong Li \\
     \texttt{cong.lea@hotmail.com} \\
}

\begin{document}
\maketitle

\begin{abstract}
Model-based reinforcement learning attempts to use an available or learned model to improve the data efficiency of reinforcement learning.
This work proposes a one-step lookback approach that jointly learns the deep incremental model and the policy to realize the sample-efficient continuous robotic control, wherein the control-theoretical knowledge is utilized to decrease the model learning difficulty and facilitate efficient training.
Specifically, we use one-step backward data to facilitate the deep incremental model, an alternative structured representation of the robotic evolution model, that accurately predicts the robotic movement but with low sample complexity.
This is because the formulated deep incremental model degrades the model learning difficulty into a parametric matrix learning problem, which is especially favourable to high-dimensional robotic applications.
The imagined data from the learned deep incremental model is used to supplement training data to enhance the sample efficiency.
Comparative numerical simulations on benchmark continuous robotics control problems are conducted to validate the efficiency of our proposed one-step lookback approach.
\end{abstract}


\section{Introduction}
The sampling inefficiency hinders model free reinforcement learning (MFRL) algorithms for continuous robotics control, wherein large amounts of expensive physical interactions between robots and environments are required before learning one satisfactory policy \cite{wang2019benchmarking}.
The motivation to improve the sample efficiency brings the advancement of the model based reinforcement learning (MBRL) field \cite{polydoros2017survey}.
MBRL is an iterative framework wherein an agent acts to collect data, learns the transition function with the collected data, and leverages the learned model for policy learning.
The enhanced sample efficiency is realized by introducing an available or learned model into the policy learning process for
generating additional training data \cite{sutton1991dyna,sutton1991planning}, computing analytic gradient \cite{deisenroth2011pilco,deisenroth2013gaussian}, and serving as the basis of planning for short rollouts \cite{chua2018deep,nagabandi2018neural}.
However, the model learning part would inevitably introduces additional computational load. 
Therefore, one efficient approach to learning a latent-space model for MBRL is highly demanded. 

This work utilizes control-theoretical knowledge to offer an alternative approach to parameterizing and learning the latent-space model for MBRL.
The formulated deep incremental model hugely decreases the model learning difficulty, and favours generality towards high-dimensional robots.

\section{Related Work}
\textbf{MFRL and MBRL:} 
The MFRL algorithms enjoy generality but suffer sample inefficiency. 
MBRL relieves the dilemma between generality and efficiency mentioned above via learning a model and using the learned model to enhance sample efficiency. 
Regarding MBRL, the core problems are how to efficiently learn a latent-space model and use the learned model for policy optimization. 

\textbf{Model learning methods:} 
The fundamental problem of MBRL is how to efficiently learn a robotics evolution model to boost the data efficiency of the policy learning process. 
Current works often parameterize the robotics evolution model in forms such as local linear time-varying system \cite{levine2015learning}, Gaussian process parameterized model \cite{deisenroth2013gaussian}, and neural network parameterized model \cite{janner2019trust}. 
There exist two main branches for learning the associated parameters of the parameterized model. One is the model-accuracy-oriented method \cite{janner2019trust} and the other is the performance-oriented approach \cite{eysenbach2022mismatched}. 
The widely utilized model-accuracy-oriented methods attempt to learn an accurate model that precisely predicts the robotic movement, in the same logic as the system identification in the control field.
However, the performance-oriented approach argues that an accurate learned model does not necessarily lead to a good performance \cite{eysenbach2022mismatched}. 
The above-mentioned objective mismatch problem is solved by learning a model related to high-performance policy, rather than the one that accurately describes the robotics movement \cite{eysenbach2022mismatched}. 
This work proposes an alternative approach to parameterize the model using the one-step backward (OSBK) data. The resulting deep incremental model serves as a different prediction modality in the MBRL field. 
The formulated deep incremental model degrades the model learning problem into an easier parametric matrix learning problem, which is especially favourable for high-dimensional robots. We follow the  model-accuracy-oriented methods to train the deep incremental model offline. 

\textbf{Model usage and policy learning:} 
When the learned model is ready, the subsequent problem is how to efficiently use the learned model to improve the learning efficiency.
The learned model is usually used in a Dyna-style \cite{sutton1991dyna,sutton1991planning} where the learned model is used to generate imagined data for the policy learning process. Note that the imagined data is generated without interacting with the environment.
Besides, the learned model is used to compute explicit gradient information \cite{deisenroth2011pilco,deisenroth2013gaussian}, which offers inferences for policy learning.
Additionally, the works \cite{chua2018deep,nagabandi2018neural} use the learned model to conduct planning for short rollouts.
We follow the Dyna-style to get additional training data for the policy learning process.

\section{Background Material}
This paper considers a Markov decision process (MDP) represented by $\mathcal{M} = \left(\mathcal{S},\mathcal{A},f, r, \gamma \right)$,
where $\mathcal{S} \in \mathbb{R}^n$, $\mathcal{A} \in \mathbb{R}^m$ are the state and action spaces, respectively;
$r: \mathcal{S} \times \mathcal{A} \to \mathbb{R}$ is the reward function;
$\gamma \in \left( 0, 1 \right)$ is the discount factor.
Denoting $s_t \in \mathcal{S}$ and $a_t \in \mathcal{A}$ as the robotic state and action.
The robotic movement would be described by the transition function $f: \mathbb{R}^n  \times \mathbb{R}^m \to \mathbb{R}^n$ that follows 
\begin{equation} \label{eq org dynamics}
    s_{t+1} = f (s_t , a_t).
\end{equation}
The continuous function $f$ is often unknown.
The robot in the state $s_t$ interacts with its surrounding environment by applying the action $a_t$ according to the policy 
$\pi(a|s)$, and evolves into the next state $s_{t+1}$ and obtains the reward $r(s,a)$.
In the following, we show how to learn a latent-space model $\hat{f}$ in Section \ref{sec Incremental Model Learning} and how to use the learned model $\hat{f}$ to improve the learning efficiency in Section \ref{sec Policy Learning with Learned Model}.

\section{Deep Incremental Model} \label{sec Incremental Model Learning}
This section first utilises the OSBK data and the control-theoretical knowledge to transform the nonlinear transfer function \eqref{eq org dynamics} into an equivalent parameterized linear form, whose parameters are then learned from input-out data in an offline way

\subsection{Incremental Model Construction}
This work introduces a constant matrix $L$ to facilitate the following model parameterization.
We first rewrite the robotic evolution model \eqref{eq org dynamics} as
\begin{equation} \label{eq new dynamics}
    s_{t+1} = L a_t + h_t,
\end{equation}
where $h_t:= f(s_t , a_t)- L a_t \in \mathbb{R}^n$ embodies all unknown model knowledge. 

Then, we estimate the unknown $h_t$ as
\begin{equation} \label{eq h estimation}
    \hat{h}_t =  h_{t-1} = s_t - L a_{t-1}.
\end{equation}

Substituting \eqref{eq h estimation} into \eqref{eq new dynamics} yields the incremental evolution model
\begin{equation} \label{eq new dynamics}
    s_{t+1} = s_t + (L + \frac{h_t - h_{t-1}}{\Delta a_t}) \Delta a_t 
    = s_{t} + L_t \Delta a_t,
\end{equation}
where $\Delta a_t = a_t - a_{t-1} \in \mathbb{R}^m$ is the incremental control input,
and $L_t = L + \frac{h_t - h_{t-1}}{\Delta \pi_t} \in \mathbb{R}^{n \times m}$ is the parametric matrix to be learned. 
Note that the explicit value of $L_t$ follows 
$L_t = \left\{ 
\begin{array}{rcl}
L_{t-1}, & \Delta a_t =0  \\ 
L_{t},   & \Delta a_t \ne 0
\end{array}
\right.$ to avoid the potential singularity.

The input-output dataset $\mathcal{D} = \left\{s_{t-1}, a_{t-1}, s_t, a_t \right\}$ will be used in the subsequent subsection to learn the parametric matrix $L_t$ to ensure that the evolution of \eqref{eq new dynamics} equivalently describes the movement of \eqref{eq org dynamics}.

\begin{remark}
    The control-theoretical knowledge and the OSBK data $\left\{s_{t-1}, a_{t-1} \right\}$ are additionally utilized in this subsection to get a model structure in an incremental form. 
    This incremental form contains the inherent property of the  original nonlinear robotics evolution model \eqref{eq org dynamics}.
    The resulting incremental evolution model  \eqref{eq new dynamics} is in a linear form with a parametric matrix $L_t$ to be learned. 
    This highly degrades the model learning difficulty.
    We have validated in Section \ref{sec numerical sim} that the simple incremental evolution model \eqref{eq new dynamics} is accurate enough to serve as the predictive model for MBRL.
\end{remark}

\begin{remark}
The observed benefits of delayed data for training is consistent with the results reported in \cite{peng2018sim}.
    Previous works have empirically found the effectiveness of the delayed data for training robotic policies\cite{peng2018sim}, wherein the previous one-step backward action is appended to the original state
    to construct the augmented state vector for training.
\end{remark}

\subsection{Model Learning}
This subsection uses the approximation ability of the deep neural network (DNN) to learn the parametric matrix $L_t$ by minimizing the difference between the true and the predicted next-step values. 

This work first represents $L_t$ in \eqref{eq new dynamics} with a fully connected, multi-layer neural network. Then, the Adam gradient ascent algorithm is adopted to train the DNN on the dataset $\mathcal{D}$ via minimizing the model prediction error :
\begin{equation} \label{eq error}
    L (\theta) := \argmin_{L (\theta)} \sum_{i=1}^{n} \left\| s_{t,i} - (s_{t-1,i} + L (\theta) \Delta a_{t,i})\right\|,
\end{equation}

Finally, we get the learned incremental evolution model
\begin{equation} \label{eq learned model}
     \hat{s}_{t+1} =  s_t + L (\theta) \Delta a_t,
\end{equation}
which is the learned representation of \eqref{eq org dynamics}.  The learned deep incremental momodel is in a physics-informed form, wherein the system response within two successive steps is utilized to facilitate learning.
This departs from approaches that directly utilizes a DNN to attempt to model the mapping between inputs and outputs.
The following section will discuss how to use the learned incremental evolution model \eqref{eq learned model} to improve the learning efficiency.

\begin{remark}
In cases where the parametric matrix $L_t$ is a square matrix, for example fully actuated robot manipulators and cars,  
we could further assume that $L_t$ is a diagonal matrix to degrade the learning difficulty. 
We found in practice that this kind of simplification is reasonable (without degrading performance but improving model learning efficiency). 
Besides, the aforementioned simplification is favourable for high-dimensional robotics applications.
\end{remark}

\begin{remark}
We could learn the deep incremental model offline using precollected data, or learn the model along with the policy learning process.
For the online model learning scenario, the data is scarce to learn a satisfactory prediction model at the beginning of the model learning process. We mitigate this problem in practice by choosing an initial value for $L (\theta)$ using the prior knowledge of the evolution model \eqref{eq org dynamics} if available. 
    The chosen initial value serves as the learning starting point of the $L(\theta)$ learning process.
    For example, the prior mass matrix of the robot manipulator, although not accurate, is enough to act as the initial value of the $L (\theta)$ learning process.
\end{remark}

\section{Policy Learning with Learned Model} \label{sec Policy Learning with Learned Model}
This section presents how to use the learned model to improve learning efficiency and how to get the robotic policy.

\subsection{Model Usage}
This subsection uses the learned deep incremental model \eqref{eq learned model} to conduct one-step forward prediction. These predictions and the collected environmental data together serve as sample data for policy evaluation and improvement.
We use one-step backward data informed model for one-step forward prediction. This contributes to fully utilizing the robotic physical information to improve learning efficiency.

\subsection{Policy Optimization}
This work focuses on the continuous robotic control problem. Therefore, we choose the soft actor-critic (SAC) algorithm to solve our problem.
In SAC, the critic agent estimates 
\begin{equation} \label{eq evaluation}
     Q^{\pi}(s,a) = E_{\pi} \left[ \sum_{t=0}^{\infty} \gamma^t r(s_t, a_t)| s_0=s, a_0 = a \right],
\end{equation}
using the Bellman backup operator, 
and the actor finds the policy $\pi$ via minimizing the expected KL-divergence
\begin{equation} \label{eq KL}
     J_{\pi}(\phi,\mathcal{D}) = \mathbb{E}_{s_t \sim \mathcal{D}} \left[ D_{KL} (\pi || \exp \left\{Q^\pi - V^\pi\right\})  \right].
\end{equation}
The corresponding pseudo code is provided in Algorithm~\ref{alg IMNRL}.

\begin{remark}
Note that our learned deep incremental model \eqref{eq learned model} is agnostic to the policy learning algorithm. 
This implies the generality of the MBRL framework in this work.
The model learning and usage modules could be used together with different policy optimization algorithms.
\end{remark}

\begin{algorithm}
\caption{Deep Incremental Model-Based Reinforcement Learning}
\label{alg IMNRL}
\begin{algorithmic}[1]
\State Initialize policy $\pi$, predictive model $f(\theta)$, environment dataset $\mathcal{D}_{\text{env}}$ and model dataset $\mathcal{D}_{\text{model}}$
\For{$N$ epochs}
    \State Train $f(\theta)$ with $\mathcal{D}_{\text{env}}$ via maximum likelihood
    \For{$E$ steps}
        \State Take action  in environment according to the policy $\pi$ and add data to the $\mathcal{D}_{\text{env}}$
        \For{$M$ model rollouts}
            \State Sample uniformly from $\mathcal{D}_{\text{env}}$
            \State Perform one-step model rollout starting from the sample using policy and add to $\mathcal{D}_{\text{model}}$
        \EndFor
        \For{$G$ gradient updates}
            \State Update policy parameters on model data
        \EndFor
    \EndFor
\EndFor
\end{algorithmic}
\end{algorithm}

\section{Online Fine-tuning}
The robotic policy trained in simulators might perform undesirably on real robots due to the gap between simulators and hardware. In addition, the policy trained in simulators might not be fully executed, as the trained policy is often directly clipped to satisfy hardware constraints. The above concerns motivate us to fine-tune the incremental policy for enhanced performance online.

Applying the incremental policy $\Delta a_d$ (trained in the simulator) to real robots yields the error $e_k:= s_{k} - s_{d}$, whose evolution could be described by 
\begin{equation} \label{eq error model}
     e_{k+1} =  e_{k} +  L(\theta) \Delta a_{e,k},
\end{equation}
wherein  $\Delta a_e$ is the residual incremental policy to be learned online to refine the pretrained $\Delta a_d$.

The error evolution model \eqref{eq error model} is in a linear form, which is convenient to use model predictive control or linear quadratic regulator to design the residual incremental policy $\Delta a_{e}$ that fine-tunes the pre-trained incremental policy $\Delta a_d$ to accommodate the ever-changing environment. 

\begin{remark}
   The deep incremental model learned offline makes the online fine-tuning incremental policy possible. Here the tools from the learning and control fields collaborate to offer a robotic policy with enhanced performance.   
\end{remark}

\section{Numerical Simulation} \label{sec numerical sim}

\begin{figure}[H]
\centering
\begin{subfigure}[b]{0.33\textwidth}
\centering
    \includegraphics[width=\textwidth]{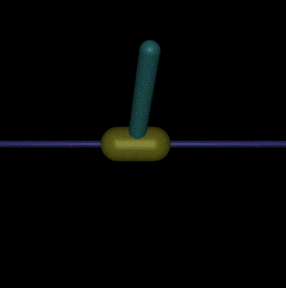}
    \caption{Inverted pendulum}
    \label{fig 1 mujoco}
\end{subfigure}
\hfill
\begin{subfigure}[b]{0.33\textwidth}
\centering
    \includegraphics[width=\textwidth]{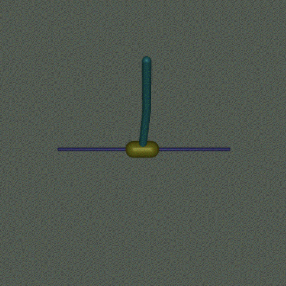}
    \caption{\centering Inverted double pendulum}
    \label{fig 2 mujoco}
\end{subfigure}
\hfill
\begin{subfigure}[b]{0.33\textwidth}
\centering
    \includegraphics[width=\textwidth]{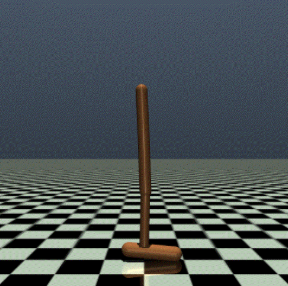}
    \caption{Hopper}
    \label{fig 3 mujoco}
\end{subfigure}
\caption{The Mujoco benchmark continuous control tasks.}
\label{fig mujoco}
\end{figure}

This section compares our method with state-of-the-art model-based (MBPO and MnM in particular) and model-free (SAC in particular) algorithms on benchmark continuous control tasks (see Figure \ref{fig mujoco}) to show the efficiency of our method.

\begin{figure}[H]
\centering
\begin{subfigure}[b]{0.33\textwidth}
\centering
    \includegraphics[width=\textwidth]{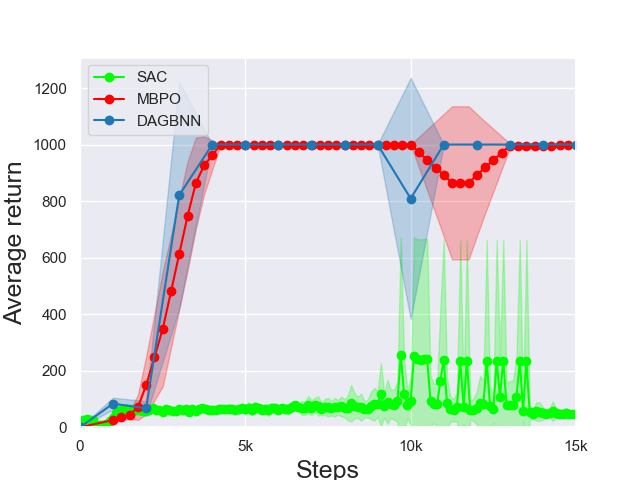}
    \caption{The learning curves of the inverted \\
    \centering{pendulum control task.}}
    \label{fig 1 result}
\end{subfigure}
\hfill
\begin{subfigure}[b]{0.33\textwidth}
\centering
    \includegraphics[width=\textwidth]{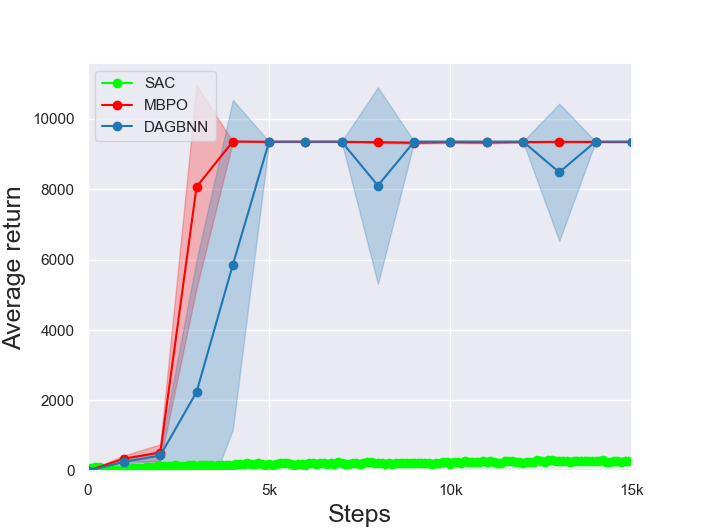}
    \caption{The learning curves of the inverted \\
    \centering{double pendulum control task.}}
    \label{fig 2 result}
\end{subfigure}
\hfill
\begin{subfigure}[b]{0.33\textwidth}
\centering
    \includegraphics[width=\textwidth]{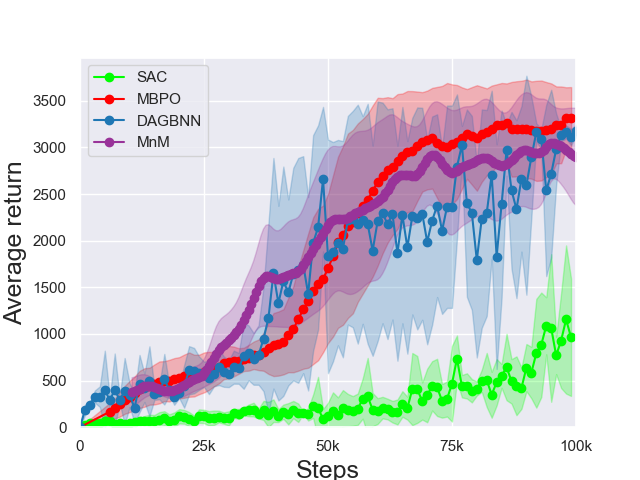}
    \caption{The learning curves of the hopper 
    \centering{control task.}}
    \label{fig 3 result}
\end{subfigure}
\caption{The learning curves of SAC, MBPO, MnM and our approach}
\label{fig results}
\end{figure}

The learning curves for all methods are presented in Fig.~\ref{fig results}. 
The comparison of our approach (which also uses SAC for policy optimization) and the baseline SAC algorithm would show the benefit of incorporating a model into the learning process. The results presented in Figure \ref{fig results} show the performance enhancement brought by the learned incremental evolution model. Our approach learns substantially faster than the model-free SAC algorithm.

The MBPO work inspires this work. This work follows the MBPO work's way to use the learned model and to optimize a policy.
Through  comparison with this milestone work, we check whether our modelling technique is more efficient than other modelling techniques. 
The comparison results presented in Figure \ref{fig results}  show that our approach achieves comparable performance with the MBPO. However, the modelling technique in our approach holds the potential to apply to high-dimensional robots, while the Gaussian process modelling technique used in MBPO is restricted to the low-dimensional domain.

The MnM work learns the latent model from a performance-oriented perspective. We compare with the MnM work to check which perspective is the right perspective to learn the model.
The simulation presented in Figure \ref{fig 3 result} shows that our approach realizes competing performance with the MnM work. Which kind of perspective is better for MBRL remains to be further explored.

\section{Discussion}
This work jointly learns and improves the model and policy from environmental interactions.
The control-theoretical knowledge and the OSBK data represent the robotic evolution model as one linear incremental form that contributes to efficient model learning. 
The learned deep incremental model serves as the prediction model in MBRL and improves the sampling efficiency. 
The formulated deep incremental model serves as one promising alternative modelling technique in MBRL.
The comparative numerical simulations on benchmark continuous robotics control tasks validate the efficiency of our approach.
\newline

\textbf{Limitations} 
The limitation is that how much nonlinearity the deep incremental model could be addressed remains to be clarified. Besides, the deep incremental model is hard to represent discontinuous dynamic systems.

\section{Future Works}
The utilized control-theoretical knowledge benefits RL with enhanced sample efficiency by offering an explicit control-oriented deep incremental model, rather than just a black-box input-output mapping.
Given stability analysis and safety checks both require one mathematical form of robotic evolutions, the explicit learned model offers us avenues to address safety and stability concerns, which remain to be further explored. 
This work uses the learned model to generate additional training data. This is one of the ways to utilize the learned model. 
It is interesting to explore the different roles of the learned model in the learning process. 
For example, the learned incremental evolution model serves as a basis for planning.
Besides, it is also interesting to investigate the performance of utilizing the deep incremental model in an ensemble way.

\bibliographystyle{unsrt}  
\bibliography{references} 

\appendix

\end{document}